\documentclass{article}
\usepackage{spconf,amsmath,graphicx,hyperref,amssymb,amsfonts}
\usepackage{cite}
\usepackage{booktabs}
\usepackage{balance}
\usepackage{multirow}
\usepackage[table]{xcolor}
\usepackage{graphicx}
\usepackage{textcomp}
\usepackage{xcolor}
\usepackage{microtype}

\newtheorem{remark}{Remark}
\title{GraTED: Graph--Tensor EEG Decoding for Passenger Hazard Perception in Autonomous Vehicles}

\name{
Yuxuan Gu$^{1}$\sthanks{Equal Contribution},
Yingkai Yang$^{1}$\footnotemark[1]\sthanks{Corresponding Author},
Qiyu Rao$^{1}$,
Tania Stathaki$^{1}$,
Danilo Mandic$^{1}$
}

\address{
$^{1}$ Department of Electrical and Electronic Engineering,
Imperial College London, United Kingdom\\
}

\begin{document}
\ninept
\maketitle

\begin{abstract}
Passenger electroencephalography (EEG) can reveal road risk through anticipatory responses before hazardous events and neural reactions around their onset. However, existing EEG decoding methods often map physiologically meaningful arrangement of electrodes onto artificial 2D grids and thus flatten the higher-order time--feature representations into matrices, obscuring meaningful structure in the data. To preserve and exploit the spatial and time--feature structure of EEG recordings, we propose the \textbf{\underline{G}}raph-\textbf{\underline{T}}ensor \textbf{\underline{E}}EG \textbf{\underline{D}}ecoding Model (\textbf{GraTED}) for passenger risk prediction (RP) and danger identification (DI) in autonomous-vehicle scenarios. GraTED models the electrode arrangement as an undirected graph based on their scalp locations, using graph convolution to capture spatial dependencies and low-rank tensor classification to preserve the multidimensional time--feature structure, while controlling model complexity. Evaluated on the PassengerEEG dataset, comprising 11 participants and 45 hours of recordings, GraTED achieves the highest balanced accuracy, compared to the state-of-the-art, across all three evaluation settings, while requiring the fewest learnable parameters among seven baseline methods.
\end{abstract}


\section{Introduction}

Autonomous vehicles (AVs) have the potential to reduce the burden of vehicle control while improving mobility and reducing accidents associated with human error. However, their safe deployment remains challenging because vehicle behavior may become less reliable in rare or complex situations, particularly when such situations are insufficiently represented in training data~\cite{yang2026eeg, Yu2024}. Moreover, objective vehicle performance does not necessarily reflect passengers' perceived risk, and mismatches between the two may undermine trust in autonomous driving. This motivates the use of complementary human-centered information. In this work, we leverage brain-computer interfaces (BCIs) to capture passengers' risk perception from electroencephalography (EEG) signals, providing an additional source of information for safer and more human-aligned AV behavior.

EEG is well-suited for capturing such responses due to its portability and high temporal resolution~\cite{di2019eeg}. Prior work reveals that passenger EEG encodes anticipatory activity before hazards and stimulus-related responses near their onset~\cite{belcher, xiaofeieeg, nguyen2019detection, yang2026eeg}, motivating two complementary tasks in the passenger-centered evaluation framework~\cite{xiaofeieeg, yang2026eeg}: Risk Prediction (RP), which uses pre-event EEG to assess potential risk, and Danger Identification (DI), which detects actual danger around event onset. Risk-aware Sequential Labeling (RSL) further enables RP-acquired knowledge to inform DI.

Existing approaches, particularly the prior 3D-CRNN~\cite{yang2026eeg}, demonstrate the value of joint spatiotemporal EEG modeling. However, the 3D-CRNN has two structural limitations. First, it maps the 54 electrodes onto a \(6 \times 9\) grid for convolution. While computationally convenient, this representation distorts the irregular 10--20 scalp geometry~\cite{Klem1999TheTE} and imposes an artificial regular neighborhood structure. Second, its GRU-based fusion treats the structured time--feature representation as a sequence, without explicitly modeling higher-order interactions between the time and feature modes. These limitations motivate the following research question:

\begin{itemize}
\item \textit{Can passenger EEG be modeled in a parameter-efficient manner, while preserving domain-specific electrode geometry and thus capturing higher-order interactions between time and feature modes for accurate hazard-perception decoding?}
\end{itemize}

\begin{figure*}[ht]
    \centering
    \includegraphics[width=\textwidth]{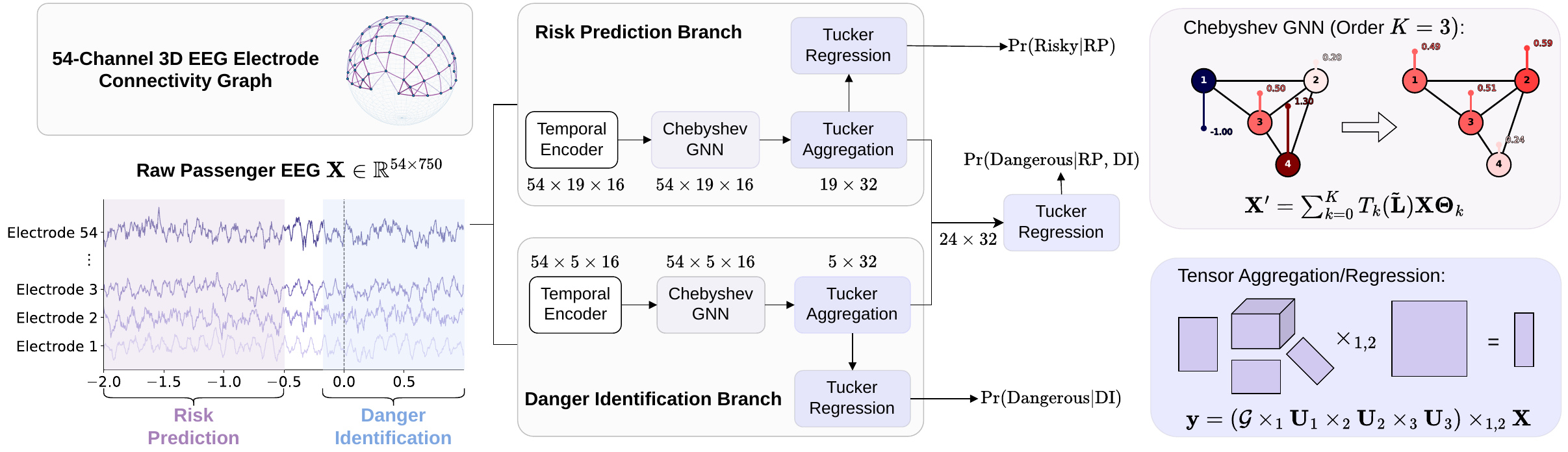}
    \caption{The GraTED architecture, consisting of risk prediction (RP) and danger identification (DI) branches.}
    \label{fig:architecture}
\end{figure*}

To this end, we propose \textbf{\underline{G}}raph-\textbf{\underline{T}}ensor \textbf{\underline{E}}EG \textbf{\underline{D}}ecoding Model (\textbf{GraTED}), a parameter-efficient framework for passenger risk prediction and danger identification in AV scenarios. Instead of imposing an artificial rectangular electrode layout, GraTED represents EEG electrodes as nodes of a graph constructed from their scalp coordinates. Chebyshev graph convolutions propagate information among neighboring electrodes without requiring explicit eigendecomposition. The resulting time-feature representation is then processed directly by low-rank tensor classification (without flattening modes), thus preserving its multidimensional structure while natively modeling interactions across temporal, feature, and class axes. Such structured parameterization provides a useful inductive bias for learning cross-mode dependencies in high-dimensional EEG data with limited sample sizes. 

% We conduct experiments on PassengerEEG~\cite{passengereeg}, following the RP–DI evaluation framework established in prior work~\cite{yang2026eeg}, to assess the accuracy and parameter efficiency to GraTED.

In summary, our main contributions are threefold: 
\begin{itemize}
    \item We introduce an undirected electrode graph based on scalp coordinates, preserving realistic irregular spatial relationships among electrodes instead of imposing an artificial regular grid.
    
    \item We develop GraTED, a graph-tensor decoding architecture that combines Chebyshev graph filtering with tensor classification, preserving both spatial topology and the structure across time and feature modes of the learned representation.

    \item We evaluate our model against seven state-of-the-art baselines on PassengerEEG dataset, demonstrating that GraTED achieves the highest balanced accuracy in all settings, while using significantly fewer trainable parameters.
\end{itemize}

\section{Related Work}\label{sec:related_work}
\subsection{EEG decoding}
Existing EEG decoding approaches encompass CNN-based models (ShallowConvNet, DeepConvNet, and EEGNet)~\cite{schirrmeister,lawhern}, transformer-based CTNet~\cite{zhao2024ctnet}, spatiotemporal DSC-ConvLSTM~\cite{li2022attention} and 3D-CRNN~\cite{yang2026eeg}, and non-neural XGB-DIM~\cite{li2023assembling}, yet none fully respects electrode topology or higher-order spatiotemporal dependencies, restricting their performance in EEG decoding tasks.

\subsection{Graphs and Tensors} Graph neural networks (GNNs)~\cite{defferrard2016convolutional,kipf2016semi} naturally model irregular data by representing entities as nodes and their relationships as edges, making them well suited to EEG electrode layouts. Tensor methods~\cite{gu2026tera,gu2026temper} preserves multi-dimensional structure and uses low-rank decompositions~\cite{gu2025tensorllm} to control model complexity. GraTED combines both, using graph convolution to capture electrode relationships and tensor classification to model time--feature interactions.

\section{Methodology}
\subsection{Preliminaries}
\subsubsection{Chebyshev Graph Convolution}
Let $\mathcal{G}=(\mathcal{V},\mathcal{E},\mathbf{A})$ denote a weighted
graph with $N$ nodes and adjacency matrix, $\mathbf{A}\in\mathbb{R}^{N\times N}$. After adding self-loops, we define
the symmetrically normalized adjacency matrix, $\mathbf{A}$, and the corresponding graph
Laplacian, $\mathbf{L}$, as
\begin{equation}
\widehat{\mathbf{A}}
=
\mathbf{D}^{-1/2}(\mathbf{A}+\mathbf{I})\mathbf{D}^{-1/2},
\qquad
\mathbf{L}
=
\mathbf{I}-\widehat{\mathbf{A}},
\label{eq:normalized_adjacency}
\end{equation}
where the degree
$D_{ii}=\sum_j(A_{ij}+I_{ij})$.
The Laplacian is next rescaled as
\begin{equation}
\widetilde{\mathbf{L}}
=
\frac{2}{\lambda_{\max}}\mathbf{L}-\mathbf{I},
\label{eq:scaled_laplacian}
\end{equation}
where $\lambda_{\max}$ is the largest eigenvalue of $\mathbf{L}$. This maps the eigenvalues of $\mathbf{L}$ to $[-1,1]$, with Chebyshev polynomials as well behaved and providing a stable approximation of spectral graph filters.

Chebyshev graph convolution~\cite{defferrard2016convolutional} approximates
a spectral graph filter using Chebyshev polynomials of
$\widetilde{\mathbf{L}}$. These polynomials are defined recursively by
\begin{equation}
T_0(\widetilde{\mathbf{L}})=\mathbf{I},
\quad
T_1(\widetilde{\mathbf{L}})=\widetilde{\mathbf{L}},
\quad
T_q(\widetilde{\mathbf{L}})
=
2\widetilde{\mathbf{L}}T_{q-1}(\widetilde{\mathbf{L}})
-
T_{q-2}(\widetilde{\mathbf{L}}).
\end{equation}
Given node features
$\mathbf{X}\in\mathbb{R}^{N\times F_{\mathrm{in}}}$, a Chebyshev
convolution of order $Q-1$ is defined as, 
\begin{equation}
\mathbf{Y}
=
\sum_{q=0}^{Q-1}
T_q(\widetilde{\mathbf{L}})
\mathbf{X}\mathbf{\Theta}_q,
\qquad
\mathbf{\Theta}_q
\in
\mathbb{R}^{F_{\mathrm{in}}\times F_{\mathrm{out}}},
\label{eq:cheb_convolution}
\end{equation}
where $\mathbf{\Theta}_q$ denotes the learnable parameters for the
$q$-th polynomial order. The resulting filter aggregates information from
local graph neighborhoods without requiring an explicit eigendecomposition.
We refer the reader to~\cite{defferrard2016convolutional} for further details.

\subsubsection{Tensor Operations and Tucker Decomposition}
A tensor $\mathcal{A}\in\mathbb{R}^{I_1\times\cdots\times I_D}$ is a
$D$-way array~\cite{cichocki2015tensor,kolda2009tensor}. We use
$a$, $\mathbf{a}$, $\mathbf{A}$, and $\mathcal{A}$ to denote scalars,
vectors, matrices, and tensors, respectively. The mode-$n$ product of
$\mathcal{A}$ with
$\mathbf{U}\in\mathbb{R}^{J\times I_n}$ is denoted by
$\mathcal{B}=\mathcal{A}\times_n\mathbf{U}$. Its element-wise definition is
\begin{equation}
B_{i_1\ldots i_{n-1}j i_{n+1}\ldots i_D}
=
\sum_{i_n=1}^{I_n}
A_{i_1\ldots i_n\ldots i_D}U_{j i_n}.
\label{eq:mode_contraction}
\end{equation}

We additionally use $\times_{p,q}$ to denote simultaneous contraction
over two tensor modes. For
$\mathcal{A}\in\mathbb{R}^{I_1\times\cdots\times I_D}$ and
$\mathbf{B}\in\mathbb{R}^{I_p\times I_q}$, with $p<q$, the contraction
$\mathcal{C}=\mathcal{A}\times_{p,q}\mathbf{B}$ is defined as
\begin{equation}
C_{i_1\ldots i_{p-1}i_{p+1}\ldots i_{q-1}i_{q+1}\ldots i_D}
=
\sum_{i_p=1}^{I_p}
\sum_{i_q=1}^{I_q}
A_{i_1\ldots i_D}B_{i_p i_q}.
\label{eq:multi_mode_contraction}
\end{equation}
Thus, modes $p$ and $q$ are contracted, while the remaining modes of
$\mathcal{A}$ are retained. Moreover, the Tucker decomposition~\cite{kolda2009tensor} represents a $D$-way tensor $\mathcal{W}\in\mathbb{R}^{I_1\times\cdots\times I_D}$ as
\begin{equation}
\mathcal{W}
=
\mathcal{G}
\times_1 \mathbf{U}^{(1)}
\times_2 \mathbf{U}^{(2)}
\times_3
\cdots
\times_D \mathbf{U}^{(D)},
\label{eq:tucker_decomposition}
\end{equation}
where
$\mathcal{G}\in\mathbb{R}^{R_1\times\cdots\times R_D}$
is the core tensor and
$\mathbf{U}^{(d)}\in\mathbb{R}^{I_d\times R_d}$
is the factor matrix associated with mode $d$.
The tuple $(R_1,\ldots,R_D)$ defines the Tucker rank.
When $\mathcal{W}$ is used as a classification coefficient tensor,
this low-rank parameterization captures interactions among its mode-specific dimensions,
while reducing the number of trainable parameters and helping mitigate
the risk of overfitting.

\subsection{Problem Formulation and Overview}
Passenger EEG is recorded from $N=54$ scalp electrodes at 250~Hz during
simulated autonomous-driving scenarios and aligned to the onset of each
traffic event. Each trial is represented as
$\mathbf{X}\in\mathbb{R}^{N\times S}$ with $S=750$ samples over
$[-2,1]$~s relative to event onset. Risk Prediction (RP) uses the pre-event
window, $\mathbf{X}_{\mathrm{RP}}\in\mathbb{R}^{54\times375}$, from
$[-2,-0.5]$~s, whereas Danger Identification (DI) uses the event-centered
window
$\mathbf{X}_{\mathrm{DI}}\in\mathbb{R}^{54\times300}$ from
$[-0.2,1]$~s. GraTED processes both windows using temporal encoding and
graph-based spatial modeling, followed by a low-rank tensor classifier. For DI with RSL, the RP
and DI representations are further combined so that pre-event risk
information can inform danger identification.

\subsection{Electrode Graph Construction}

Each EEG electrode is represented as a graph node with normalized 3D scalp
coordinate $\mathbf{p}_i\in\mathbb{R}^{3}$ from the standard 10--20 montage.
The geodesic distance between electrodes $i$ and $j$ is $d_{ij}
=
\arccos\left(\mathbf{p}_i^{\mathsf T}\mathbf{p}_j\right).$
In our design, each node is connected to its $k=4$ nearest neighbors according to the geodesic distance, and the resulting $k$-nearest-neighbor graph is symmetrized by including an undirected edge $(i,j)$ if either $i$ selects $j$ or $j$ selects $i$ as a nearest neighbor. The edge weights are defined based on an exponential kernel, as
\begin{equation}
A_{ij}
=
\begin{cases}
\exp\left(-d_{ij}^{2}/2\sigma^{2}\right),
& j\in\mathcal{N}_k(i)\ \text{or}\ i\in\mathcal{N}_k(j),\\
0, & \text{otherwise},
\end{cases}
\end{equation}
where $\sigma$ is the median pairwise electrode distance. The resulting
adjacency matrix is normalized with self-loops according to Eq.
(\ref{eq:normalized_adjacency}). This graph reflects physical
proximity between scalp electrodes, without imposing an artificial
rectangular layout.

\subsection{Temporal and Graph Feature Encoding}
\begin{table*}[t]
\caption{Performance and model-size comparison (mean $\pm$ standard deviation across 11 participants). Params. denotes the trainable parameters for each setting. The best results in each metric are bold.}
\label{tab:results}
\centering
\setlength{\tabcolsep}{3.5pt}
\begin{tabular}{p{1.5cm} p{3.0cm} p{1.5cm} p{2.4cm} p{2.4cm} p{2.4cm} p{2.7cm}}
\toprule
Setting & Model & Params. $\downarrow$ & Precision $\uparrow$ & Recall $\uparrow$ & F1-Score $\uparrow$ & Balanced Accuracy $\uparrow$ \\
\midrule
\multirow{8}{*}{RP}
 & EEGNet & 6,762 & 0.916 $\pm$ 0.068 & 0.909 $\pm$ 0.047 & 0.911 $\pm$ 0.051 & 0.917 $\pm$ 0.049 \\
 & ShallowConvNet & 89,002 & 0.963 $\pm$ 0.036 & \textbf{0.918 $\pm$ 0.045} & 0.939 $\pm$ 0.036 & 0.943 $\pm$ 0.033 \\
 & DeepConvNet & 298,852 & 0.951 $\pm$ 0.040 & 0.918 $\pm$ 0.042 & 0.934 $\pm$ 0.037 & 0.938 $\pm$ 0.035 \\
 & XGB-DIM & 17,850 & 0.635 $\pm$ 0.125 & 0.427 $\pm$ 0.207 & 0.474 $\pm$ 0.170 & 0.596 $\pm$ 0.068 \\
 & CTNet & 21,890 & 0.923 $\pm$ 0.074 & 0.857 $\pm$ 0.172 & 0.881 $\pm$ 0.138 & 0.901 $\pm$ 0.096 \\
 & DSC-ConvLSTM & 7,232 & 0.935 $\pm$ 0.079 & 0.887 $\pm$ 0.070 & 0.909 $\pm$ 0.067 & 0.915 $\pm$ 0.062 \\
 & 3D-CRNN & 33,410 & 0.959 $\pm$ 0.038 & 0.903 $\pm$ 0.060 & 0.929 $\pm$ 0.045 & 0.934 $\pm$ 0.041 \\
  \rowcolor{blue!10}
 & \textbf{GraTED (Ours)} & \textbf{6,388} & \textbf{0.977 $\pm$ 0.034} & 0.913 $\pm$ 0.044 & \textbf{0.943 $\pm$ 0.028} & \textbf{0.946 $\pm$ 0.026} \\
\midrule
\multirow{8}{*}{DI w/o RSL}
 & EEGNet & 6,602 & 0.622 $\pm$ 0.068 & 0.672 $\pm$ 0.138 & 0.642 $\pm$ 0.094 & 0.749 $\pm$ 0.068 \\
 & ShallowConvNet & 88,602 & 0.638 $\pm$ 0.047 & 0.812 $\pm$ 0.081 & 0.711 $\pm$ 0.037 & 0.806 $\pm$ 0.036 \\
 & DeepConvNet & 298,452 & 0.597 $\pm$ 0.082 & 0.709 $\pm$ 0.132 & 0.645 $\pm$ 0.093 & 0.751 $\pm$ 0.073 \\
 & XGB-DIM & 17,850 & 0.350 $\pm$ 0.044 & 0.795 $\pm$ 0.109 & 0.482 $\pm$ 0.043 & 0.573 $\pm$ 0.044 \\
 & CTNet & 21,842 & 0.551 $\pm$ 0.204 & 0.640 $\pm$ 0.281 & 0.580 $\pm$ 0.231 & 0.727 $\pm$ 0.123 \\
 & DSC-ConvLSTM & 7,232 & \textbf{0.654 $\pm$ 0.068} & 0.668 $\pm$ 0.205 & 0.634 $\pm$ 0.111 & 0.749 $\pm$ 0.069 \\
 & 3D-CRNN & 42,754 & 0.636 $\pm$ 0.051 & 0.805 $\pm$ 0.076 & 0.709 $\pm$ 0.054 & 0.803 $\pm$ 0.048 \\
 \rowcolor{blue!10}
 & \textbf{GraTED (Ours)} & \textbf{5,451} & 0.653 $\pm$ 0.027 & \textbf{0.848 $\pm$ 0.053} & \textbf{0.737 $\pm$ 0.023} & \textbf{0.827 $\pm$ 0.026} \\
\midrule
\multirow{8}{*}{DI w/ RSL}
 & EEGNet & 16,790 & 0.659 $\pm$ 0.052 & 0.826 $\pm$ 0.093 & 0.730 $\pm$ 0.055 & 0.820 $\pm$ 0.048 \\
 & ShallowConvNet & 181,446 & \textbf{0.682 $\pm$ 0.048} & 0.846 $\pm$ 0.064 & 0.753 $\pm$ 0.034 & 0.837 $\pm$ 0.031 \\
 & DeepConvNet & 607,994 & 0.679 $\pm$ 0.067 & 0.846 $\pm$ 0.073 & 0.750 $\pm$ 0.046 & 0.835 $\pm$ 0.038 \\
 & XGB-DIM & 35,703 & 0.415 $\pm$ 0.144 & 0.288 $\pm$ 0.207 & 0.313 $\pm$ 0.184 & 0.571 $\pm$ 0.063 \\
 & CTNet & 45,654 & 0.648 $\pm$ 0.050 & 0.806 $\pm$ 0.175 & 0.707 $\pm$ 0.112 & 0.809 $\pm$ 0.073 \\
 & DSC-ConvLSTM & 17,186 & 0.637 $\pm$ 0.033 & 0.822 $\pm$ 0.166 & 0.709 $\pm$ 0.080 & 0.809 $\pm$ 0.069 \\
 & 3D-CRNN & 79,334 & 0.668 $\pm$ 0.032 & 0.861 $\pm$ 0.082 & 0.750 $\pm$ 0.035 & 0.838 $\pm$ 0.035 \\
  \rowcolor{blue!10}
 & \textbf{GraTED (Ours)} & \textbf{12,805} & 0.660 $\pm$ 0.015 & \textbf{0.894 $\pm$ 0.031} & \textbf{0.759 $\pm$ 0.010} & \textbf{0.848 $\pm$ 0.012} \\
\bottomrule
\end{tabular}
\end{table*}

Temporal features are first extracted independently from each electrode,
without sharing information across nodes. Let
$\mathbf{x}_{\mathrm{RP},i}
=\mathbf{X}_{\mathrm{RP}}[i,:]\in\mathbb{R}^{375}$
and
$\mathbf{x}_{\mathrm{DI},i}
=\mathbf{X}_{\mathrm{DI}}[i,:]\in\mathbb{R}^{300}$
denote the RP and DI signals of electrode $i$, respectively. We define the
temporal encoder as
\begin{equation}
\phi(\mathbf{x}_i)
=
\log\left(
\operatorname{AvgPool}
\left[
\operatorname{BN}
\left(
\operatorname{Conv}_{1\mathrm{D}}(\mathbf{x}_i)
\right)^{\odot 2}
\right]
\right),
\label{eq:temporal_encoding}
\end{equation}
where $(\cdot)^{\odot 2}$ denotes element-wise squaring. The encoder maps each RP electrode signal to
$\mathbb{R}^{19\times16}$ and each DI electrode signal to
$\mathbb{R}^{5\times16}$, where the first dimension indexes temporal positions and the second indexes learned features. Stacking the representations across all electrodes gives
$\mathcal{H}^{(0)}_{\mathrm{RP}}\in\mathbb{R}^{N\times19\times16}$ and
$\mathcal{H}^{(0)}_{\mathrm{DI}}\in\mathbb{R}^{N\times5\times16}$.

The encoded node features are then propagated over the electrode graph to
model spatial dependencies. Using the rescaled Laplacian in
Eq.~(\ref{eq:scaled_laplacian}), we set $\lambda_{\max}=2$. Since $\mathbf{L}=\mathbf{I}-\widehat{\mathbf{A}}$, Eq.~(\ref{eq:scaled_laplacian}) reduces to
$\widetilde{\mathbf{L}}=-\widehat{\mathbf{A}}$. At graph layer $\ell$, the
node--feature representation at time index $t$ is
$\mathbf{H}^{(\ell)}_{t}
=\mathcal{H}^{(\ell)}[:,t,:]\in\mathbb{R}^{N\times16}$.
The Chebyshev convolution in Eq.~(\ref{eq:cheb_convolution}) is incorporated into a residual normalization block as
\begin{equation}
\mathbf{H}^{(\ell+1)}_{t}
=
\operatorname{LN}\left(
\mathbf{H}^{(\ell)}_{t}
+
\sum_{q=0}^{Q-1}
T_q(\widetilde{\mathbf{L}})
\mathbf{H}^{(\ell)}_{t}
\mathbf{\Theta}^{(\ell)}_q
\right),
\label{eq:cheb_layer}
\end{equation}
where $\mathbf{\Theta}^{(\ell)}_q\in\mathbb{R}^{16\times16}$ denotes the
learnable graph-filter parameters and $\operatorname{LN}(\cdot)$ denotes
layer normalization. We set $Q=3$ and use two graph layers to capture local multi-hop
interactions between electrodes. After graph propagation, each
$\mathbf{H}^{(L)}_t\in\mathbb{R}^{N\times16}$
is mapped to a $32$-D graph-level representation using
Tucker-parameterized aggregation. Following Eq.~(\ref{eq:tucker_decomposition}), we
define the coefficient tensor
$\mathcal{W}^{(\mathrm{agg})}\in\mathbb{R}^{32\times N\times16}$,
whose modes correspond to output feature, electrode, and graph features, respectively.
The associated factor matrices are
$\mathbf{U}^{(o)}\in\mathbb{R}^{32\times R_o}$,
$\mathbf{U}^{(n)}\in\mathbb{R}^{N\times R_n}$, and
$\mathbf{U}^{(h)}\in\mathbb{R}^{16\times R_h}$.
We use Tucker ranks $(R_o,R_n,R_h)=(8,24,8)$. At each time index, the graph-level representation $\mathbf{z}_t\in\mathbb{R}^{32}$ is obtained by contracting the electrode
and graph-feature modes of $\mathcal{W}^{(\mathrm{agg})}$ with
$\mathbf{H}^{(L)}_t$, as
\begin{equation}
\mathbf{z}_t
=
\sigma\left(
\mathcal{W}^{(\mathrm{agg})}
\times_{2,3}
\mathbf{H}^{(L)}_t
+
\mathbf{b}^{(r)}
\right),
\label{eq:graph_aggregation}
\end{equation}
where $\mathbf{b}^{(r)}\in\mathbb{R}^{32}$ and
$\sigma(\cdot)$ denotes the ELU activation function~\cite{clevert2015fast}. Collecting these representations over time gives
$\mathbf{Z}=[\mathbf{z}_1,\ldots,\mathbf{z}_T]^T
\in\mathbb{R}^{T\times32}$.
Specifically, $\mathbf{Z}_{\mathrm{RP}}\in\mathbb{R}^{19\times32}$ for RP branch and
$\mathbf{Z}_{\mathrm{DI}}\in\mathbb{R}^{5\times32}$ for DI branch.

\subsection{Tucker-Parameterized Tensor Classification}

GraTED uses separate tensor classifiers for three tasks: RP, DI without RSL, and DI
with RSL. The first two classifiers operate directly on
$\mathbf{Z}_{\mathrm{RP}}\in\mathbb{R}^{19\times32}$ and
$\mathbf{Z}_{\mathrm{DI}}\in\mathbb{R}^{5\times32}$, respectively.
For DI with RSL, the two representations are first concatenated along the temporal dimension as
$\mathbf{Z}_{\mathrm{RSL}}
=
[\mathbf{Z}_{\mathrm{RP}}\Vert\mathbf{Z}_{\mathrm{DI}}]
\in\mathbb{R}^{24\times32}$.
For each classifier, the coefficient tensor is
$\mathcal{W}^{(\mathrm{cls})}\in\mathbb{R}^{T\times32\times C}$,
where $C=2$ and $T=19$, $5$, and $24$ for RP, DI without RSL, and DI
with RSL, respectively. Each coefficient tensor is parameterized using
the Tucker decomposition in Eq.~(\ref{eq:tucker_decomposition}) over
the time, learned-feature, and prediction class modes. Given the corresponding representation
$\mathbf{Z}\in\mathbb{R}^{T\times32}$, the logits and class probabilities are
\begin{equation}
\mathbf{s}
=
\mathcal{W}^{(\mathrm{cls})}
\times_{1,2}\mathbf{Z}
+
\mathbf{b},
\qquad
\widehat{\mathbf{y}}
=
\operatorname{softmax}(\mathbf{s}),
\label{eq:tucker_classifier}
\end{equation}
where
$\mathbf{s},\widehat{\mathbf{y}},\mathbf{b}\in\mathbb{R}^{C}$.
The RP classifier predicts non-risky versus risky, while both DI
classifiers predict non-dangerous versus dangerous. The predicted class is
$\widehat{c}=\arg\max_c\widehat{y}_c$.

\begin{remark}
    An additional benefit of the Tucker parameterization is parameter efficiency, which naturally arises from its low-rank structure. For $\mathcal{W}^{(\mathrm{cls})} \in \mathbb{R}^{T \times F \times C}$, a dense parameterization would require $TFC+C$ trainable parameters, whereas a Tucker parameterization with ranks $(R_t,R_f,R_c)$ only requires $R_tR_fR_c + TR_t + FR_f + CR_c + C$, with Tucker ranks chosen much smaller than corresponding mode dimensions. The resulting
    low-rank structure provides an inductive bias towards compact, structured interactions across the temporal, feature, and output modes, while simultaneously reducing the number of trainable
    parameters. This compact parameterization also acts as an implicit regularizer, helping reduce the risk of overfitting in limited-data EEG settings.
\end{remark}

\section{Experiments}

\subsection{Experimental Setup}
We conducted experiments on PassengerEEG~\cite{passengereeg}, a preprocessed passenger EEG dataset comprising trials from 11 participants across 15 traffic scenarios, and follow the experimental setup in \cite{yang2026eeg}. Each trial belongs to one of three source categories: safe, low-risk, or high-risk. For RP, safe trials are labeled
non-risky, whereas low-risk and high-risk trials are labeled risky.
For DI, safe and low-risk trials are labeled non-dangerous, whereas high-risk trials are labeled dangerous. The same labeling scheme is used for DI with and without RSL.

For each participant, 60\% of data was used for training
and 40\% for testing, with 30\% of the training data reserved for
validation. All models were trained using AdamW optimizer \cite{loshchilov2017decoupled} with a batch size of 1024 for at most 60 epochs in each task, with early stopping based on validation balanced
accuracy (BA). Results were averaged across participants. Precision, recall, and F1-score were computed for the positive class (risky for RP and dangerous for DI), whereas BA averages recall across both classes and is therefore used as the primary metric. We compared GraTED against the seven baselines introduced in Section~\ref{sec:related_work}, using the same data splits and training setup for all models.

\subsection{Numerical Results}

Table~\ref{tab:results} shows that GraTED consistently achieves the highest
balanced accuracy across all three evaluation settings. For RP, the BA of GraTED (0.946) slightly exceeds that of ShallowConvNet (0.943), and GraTED also exhibits comparatively low
variability across participants, indicating stable RP performance.

The advantage of GraTED is more pronounced for DI without RSL, where its
BA reaches 0.827, improving on ShallowConvNet and 3D-CRNN by 2.1 and
2.4\%, respectively. Relative to these two baselines, improvements are also observed in
precision, recall, and F1-score, indicating consistent improvements across multiple classification metrics. These results demonstrate the effectiveness of GraTED for extracting danger-related information from the event-centered EEG window.

Incorporating the RP representation through RSL further improves DI
performance, increasing GraTED's BA from 0.827 to 0.848 and its F1-score
from 0.737 to 0.759. GraTED therefore outperforms the strongest
baseline under this setting, 3D-CRNN, by 1.2\%. This improvement indicates that pre-event risk information provides complementary context to the event-centered EEG response, supporting the joint use of RP and DI representations for danger identification.

\subsection{Parameter Efficiency}

GraTED remains parameter-efficient across all evaluation settings, using 6,388 parameters for RP, 5,451 for DI without RSL, and 12,805 for DI with RSL. Compared with 3D-CRNN, these correspond to reductions of 80.9\%, 87.3\%, and 83.9\%, respectively, while GraTED achieves higher BA in all three settings. The Tucker-parameterized classifiers contain only 1,028 parameters for RP, 283 for standalone DI, and 966 for DI with RSL. For DI with RSL, the Tucker classification module therefore uses 69.5\% fewer parameters than the 3,170-parameter GRU fusion module employed by 3D-CRNN. These results demonstrate that the low-rank tensor formulation yields substantial parameter savings without compromising decoding performance.

\section{Conclusion}

We have proposed GraTED, a graph--tensor framework for passenger EEG decoding that combines electrode-graph modeling with tensor classification, thus preserving both scalp geometry and the time--feature structure of the learned representation. Among seven baselines considered, GraTED has achieved the highest BA across three task settings, while reducing the active parameter count by 80.9\%--87.3\% relative to 3D-CRNN, the strongest baseline method. These results have highlighted the promise of graph--tensor modeling as an effective and parameter-efficient approach to passenger hazard perception. Future work will examine subject-independent generalization and real-world online evaluation.

\section{Compliance with Ethical Standards}
This is a numerical simulation study for which no ethical approval was required.

\bibliographystyle{IEEEbib}
\bibliography{refs}

\end{document}